\documentclass[sigconf]{acmart}

\usepackage{pifont}
\usepackage{multirow}
\usepackage{stfloats}
\usepackage{enumitem}
\usepackage{overpic}
\usepackage{bm}

\definecolor{greencolor}{RGB}{34,139,34}
\definecolor{lightred}{RGB}{255,100,100}

\AtBeginDocument{%
  \providecommand\BibTeX{{%
    \normalfont B\kern-0.5em{\scshape i\kern-0.25em b}\kern-0.8em\TeX}}}

\copyrightyear{2023} 
\acmYear{2023} 
\setcopyright{acmlicensed}\acmConference[MM '23]{Proceedings of the 31st
ACM International Conference on Multimedia}{October 29-November 3,
2023}{Ottawa, ON, Canada}
\acmBooktitle{Proceedings of the 31st ACM International Conference on
Multimedia (MM '23), October 29-November 3, 2023, Ottawa, ON, Canada}
\acmPrice{15.00}
\acmDOI{10.1145/3581783.3611933}
\acmISBN{979-8-4007-0108-5/23/10}

\newcommand{\ie}{\textit{i.e.}}
\newcommand{\eg}{\textit{e.g.}}
\newcommand{\etal}{\textit{et al.}}

\begin{document}

\title{FastLLVE: Real-Time Low-Light Video Enhancement with Intensity-Aware Lookup Table}


\author{Wenhao Li}
\authornote{Both authors contributed equally to this research.}
\affiliation{%
  \institution{Shanghai Jiao Tong University}
  \city{Shanghai}
  \country{China}
}
\email{wenhaoli.233.411@gmail.com}

\author{Guangyang Wu}
\authornotemark[1]
\affiliation{%
  \institution{Shanghai Jiao Tong University}
  \city{Shanghai}
  \country{China}
}
\email{wu.guang.young@gmail.com}

\author{Wenyi Wang}
\affiliation{%
 \institution{University of Electronic Science and Technology of China}
 \city{Chengdu}
 \country{China}
}
\email{wangwenyi@uestc.edu.cn}

\author{Peiran Ren}
\affiliation{%
  \institution{Alibaba Damo Academy}
  \city{Hangzhou}
  \country{China}
}
\email{peiran_r@sohu.com}

\author{Xiaohong Liu}
\authornote{Corresponding Author: xiaohongliu@sjtu.edu.cn}
\affiliation{%
  \institution{Shanghai Jiao Tong University}
  \city{Shanghai}
  \country{China}
}
\email{xiaohongliu@sjtu.edu.cn}


\renewcommand{\shortauthors}{Wenhao Li, Guangyang Wu, Wenyi Wang, Peiran Ren,\& Xiaohong Liu}


\begin{abstract}
    Low-Light Video Enhancement (LLVE) has received considerable attention in recent years. One of the critical requirements of LLVE is inter-frame brightness consistency, which is essential for maintaining the temporal coherence of the enhanced video. However, most existing single-image-based methods fail to address this issue, resulting in flickering effect that degrades the overall quality after enhancement. Moreover, 3D Convolution Neural Network (CNN)-based methods, which are designed for video to maintain inter-frame consistency, are computationally expensive, making them impractical for real-time applications.
    To address these issues, we propose an efficient pipeline named \textit{FastLLVE} that leverages the Look-Up-Table (LUT) technique to maintain inter-frame brightness consistency effectively. Specifically, we design a learnable Intensity-Aware LUT (IA-LUT) module for adaptive enhancement, which addresses the low-dynamic problem in low-light scenarios. This enables FastLLVE to perform low-latency and low-complexity enhancement operations while maintaining high-quality results.
    Experimental results on benchmark datasets demonstrate that our method achieves the State-Of-The-Art (SOTA) performance in terms of both image quality and inter-frame brightness consistency. More importantly, our FastLLVE can process 1,080p videos at $\mathit{50+}$ Frames Per Second (FPS), which is $\mathit{2 \times}$ faster than SOTA CNN-based methods in inference time, making it a promising solution for real-time applications. \textit{The code is available at https://github.com/Wenhao-Li-777/FastLLVE.}
\end{abstract}



\maketitle

\section{Introduction}
Low-Light Video Enhancement (LLVE) is a longstanding task aiming at transforming low-light videos into normal-light videos with better visibility, which has received considerable attention in recent years. In low-light conditions, videos often suffer from deteriorated texture and low contrast, leading to poor visibility and significant degradation of high-level vision tasks. Unlike traditional methods based on higher ISO and exposure that can cause noise and motion blur~\cite{cheng2016learning}, LLVE offers an effective solution to improve the visual quality of videos captured in extremely low-light conditions. Moreover, it can serve as a fundamental enhancement module for a wide range of applications, \eg, visual surveillance~\cite{yang2019coarse}, autonomous driving~\cite{li2021deep}, and unmanned aerial vehicle~\cite{samanta2018log}.

Like other typical video tasks, such as Video Frame Interpolation~\cite{shi2021video, shi2022video, wu2023accflow} and Video Super-Resolution~\cite{liu2018robust, liu2020end, liu2021exploit, shi2021learning, huang2023transmrsr, yin2023online}, LLVE also demands temporal stability. Additionally, the inherently ill-posed nature of LLVE makes it a more challenging task. As a result, although Low-Light Image Enhancement (LLIE) have demonstrated remarkable performance, recursively applying these image-based methods to video frames isn't feasible. Because it is time-consuming and may result in flickering effect in the enhanced video. As revealed in~\cite{jiang2019learning}, the flickering problem is caused by the inconsistency in brightness between adjacent frames. To address this issue, recent LLVE methods have leveraged temporal alignment~\cite{wang2021seeing} and 3D Convolution (3D-Conv)~\cite{lv2018mbllen,jiang2019learning} to establish the spatial-temporal relationship in video. They have also adopted the self-consistency~\cite{chen2019seeing,zhang2021learning} as an auxiliary loss to guide the network in maintaining brightness consistency. However, alignment-based methods, which aim to estimate the corresponding pixels between adjacent frames, are prone to errors and can lead to object distortion in the enhanced video. In contrast, 3D-Conv is capable of capturing comprehensive spatial-temporal information, but at the cost of greater computational complexity. Therefore, previous methods have found it challenging to strike a balance between efficiency and performance. To sum up, a considerate LLVE method should address the following challenging issues: 

\noindent $\bullet$ \textbf{Ill-posed problem.} In low-light videos, the low dynamic range of the color space can result in similar color inputs appearing for different target colors. This phenomenon leads to the one-to-many mapping problem which is challenging to solve in complex scenarios. To address this problem, previous methods~\cite{lv2018mbllen, jiang2019learning, wang2021seeing} have leveraged global context information and local consistency to enhance different colors. 
Despite their respective efficacy, these methods are plagued by instability with respect to color handling due to their heavy reliance on the precision and reliability of context extraction.

\noindent $\bullet$ \textbf{Brightness consistency.} Maintaining brightness consistency in the output video is crucial for achieving high perceptual quality in LLVE. However, current alignment-based method~\cite{wang2021seeing} often fails to achieve accurate alignment between adjacent frames, leading to unstable output for LLVE. Otherwise, self-consistency loss
functions~\cite{chen2019seeing, zhang2021learning} used to improve the stability of these methods are also unable to address the fundamental instability problem. This limitation hinders their ability to effectively improve their overall visual quality.
    
\noindent $\bullet$ \textbf{Efficiency.} Although 3D-Conv methods have shown significant improvement in video enhancement tasks by exploiting comprehensive spatial-temporal information~\cite{lv2018mbllen, jiang2019learning}, they are associated with heavy computational complexity. This makes them impractical for real-world applications that require real-time enhancement. 

\begin{figure}[t]
	\centering
	\includegraphics[width=\linewidth]{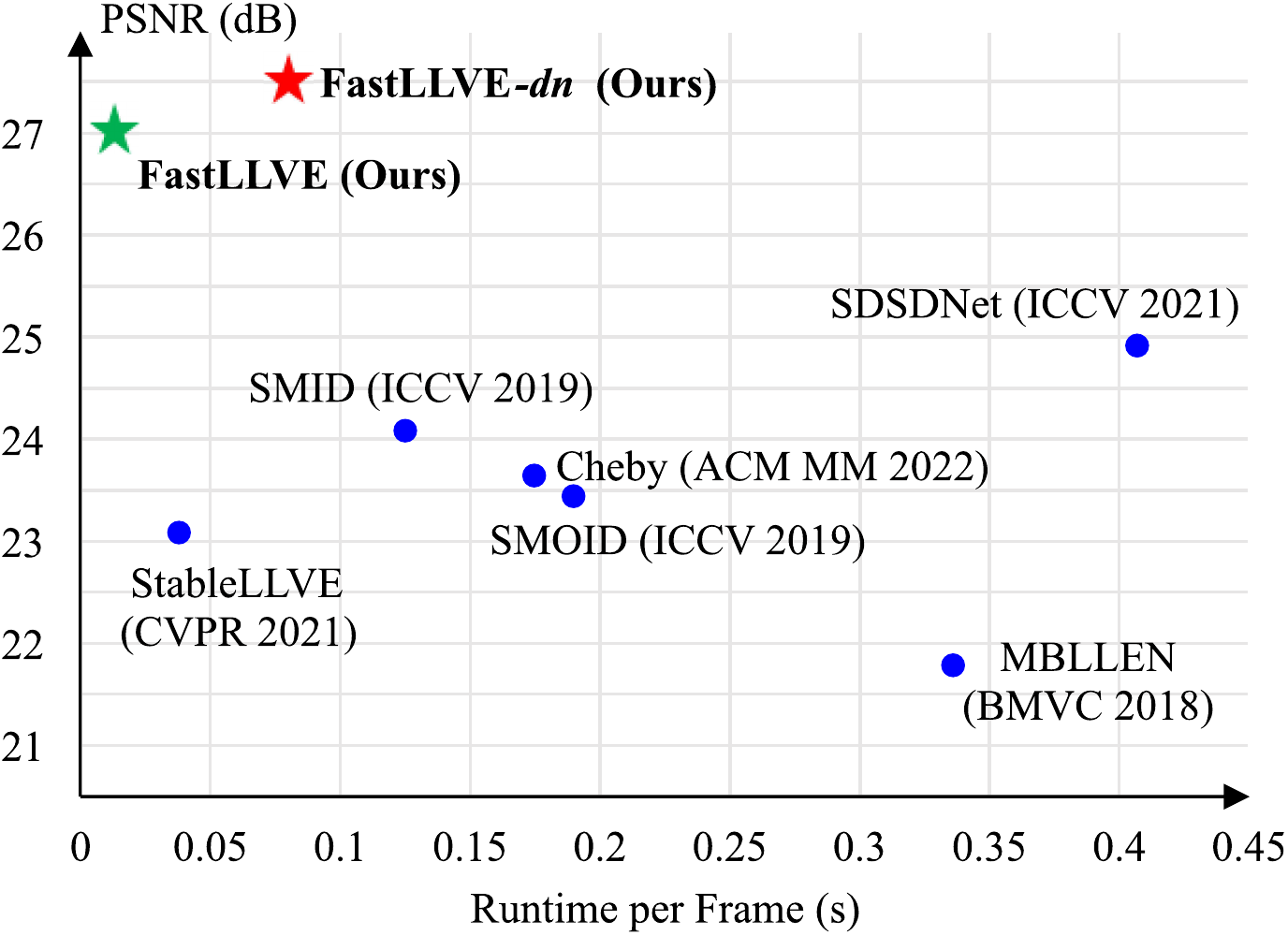}
    \caption{Comparisons of effectiveness and efficiency. Our method outperforms the current SOTA method (\ie, SDSDNet) by a large margin in terms of PSNR, and faster than the most efficient method (\ie, StableLLVE). The average PSNR is evaluated on SDSD test dataset~\cite{wang2021seeing}, and runtime is evaluated on 1,080p videos with a Nvidia RTX 3090 GPU.}
	\label{fig:teaser}
    \vspace{-4mm}
\end{figure}
To address the above issues, we propose a novel framework named FastLLVE. Our approach establishes a stable and adaptive Look-Up-Table (LUT) to enable real-time LLVE. In particular, we design an Intensity-Aware LUT (IA-LUT) to transform RGB colors from one color space to another, which can handle the one-to-many mapping issue that commonly arises in LLVE. Unlike traditional LUTs where one-to-one mapping relationships for color values are stored, our IA-LUT stores the one-to-many mapping relationships, with respect to learnable enhancement intensities for every pixels. To improve the generalization ability of our method, we follow the parameterization approach~\cite{zeng2020learning} and combine a set of basis LUTs with dynamic weights. Importantly, our approach maintains the inter-frame brightness consistency by nature, as the pixel-wise LUT-based transformation is consistent with all pixels having the same RGB values and enhancement intensity. In addition, our method is computationally efficient and suitable for real-time video enhancement. To address the issue of noise that the LUT might fail to deal with, particularly in extremely low-light conditions, we simplify a common denoising method~\cite{chen2019real} to incorporate a plug-in refinement module for denoising denoted as FastLLVE-\textit{dn}, which further improves the performance at the expense of some efficiency. It is worth noting that other denoising methods can readily replace the used one. As demonstrated in Figure~\ref{fig:teaser}, both of our two models outperform existing methods by a significant margin in terms of Peak Signal-to-Noise Ratio (PSNR), while the FastLLVE achieves the real-time processing speed of over 50 Frames Per Second (FPS).

The contributions of this paper can be summarized as follows:

$\diamond$ We propose a novel LUT-based framework, named FastLLVE, for real-time low-light video enhancement.

$\diamond$ We design a novel and lightweight Intensity-Aware LUT, which accounts for the one-to-many mapping problem in LLVE.

$\diamond$ Extensive experiments show that the FastLLVE achieves the SOTA results on benchmarks in most cases, with over 50 FPS inference speed.
\section{Related works}

\subsection{Low-light Image Enhancement}
Researches on low-light enhancement started with traditional LLIE methods including Histogram Equalization~\cite{ibrahim2007brightness, wang2007fast, nakai2013color} and Retinex theory~\cite{land1977retinex, wang2013naturalness, fu2015probabilistic, fu2016weighted}. Then deep-learning approaches~\cite{lv2018mbllen, lai2018learning, wang2019underexposed, moran2020deeplpf, yang2020fidelity, pan2022chebylighter, zhou2022lednet} have shown the great superiority on effectiveness, efficiency and generalization ability. Lv~\etal~\cite{lv2018mbllen} present a multi-branch network, which extracts rich features from different levels, to enhance low-light images via multiple subnets. Wang~\etal~\cite{wang2019underexposed} introduce intermediate illumination rather than directly learn an image-to-image mapping. Pan~\etal~\cite{pan2022chebylighter} propose a new model learning to estimate pixel-wise adjustment curves and recurrently reconstruct the output. Zhou~\etal~\cite{zhou2022lednet} specially design a network for joint low-light enhancement and deblurring.

\subsection{Low-light Video Enhancement}
LLVE, an extension of LLIE, imposes an additional requirement of brightness consistency, as outlined in~\cite{lai2018learning}. Existing LLVE methods address this challenge through three common solutions, namely 3D Convolution, Feature Alignment, and Self-consistency.
Lv~\etal~\cite{lv2018mbllen} exchange all 2D-Conv layers of their proposed LLIE network into 3D-Conv layers to achieve the processing of low-light videos. 
Jiang~\etal~\cite{jiang2019learning} train a LLVE network based on 3D U-Net. 
Instead of 3D-Conv, Wang~\etal~\cite{wang2021seeing} align adjacent frames into the middle frame for lighting enhancement and noise reduction based on Retinex theory~\cite{land1977retinex}. 
In order to improve efficiency, some methods use 2D-Conv with self-consistency as an auxiliary loss. Chen~\etal~\cite{chen2019seeing} randomly select two frames from the same low-light video to train a deep twin network, using self-consistency loss to make the network robust to noise and small changes in the scene. Rather than select similar frames, Zhang~\etal~\cite{zhang2021learning} choose to simulate adjacent frames and ground truths by warping the input image and its corresponding ground truth based on the predicted optical flow, so as to artificially synthesize similar data pairs for self-consistency loss. However, self-consistency is a weak and unstable constraint which cannot solve the fundamental problem of brightness consistency.

\begin{figure*}[t]
	\centering
	\includegraphics[width=0.98\linewidth]{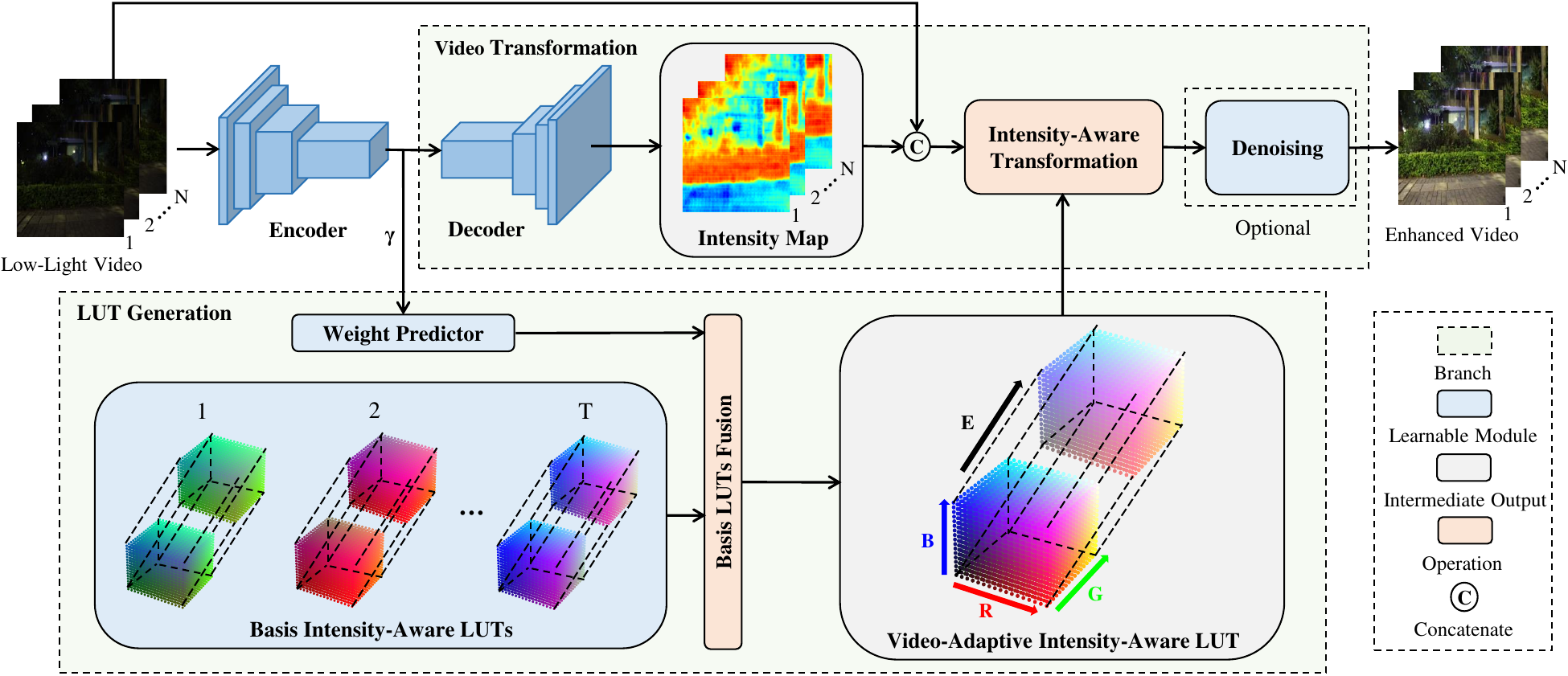}
	\caption{The architecture of the proposed network based on the designed Intensity-Aware LUT (IA-LUT). The lightweight encoder-decoder network extracts spatial and temporal features for building a video-adaptive IA-LUT and generates intensity map related to the input video. Then weight predictor utilizes the feature vector $\gamma$ from the encoder to predict weights that guide the fusion of $T$ basis IA-LUTs. Through IA-LUT transformation, the input video concatenated with intensity map transforms to the enhanced normal-light video. Finally, an optional denoising module can help the IA-LUT deal with noise.}
	\label{fig:architecture}
	\vspace{-4mm}
\end{figure*}
\subsection{LUT for Image Enhancement}
A 3D-LUT is a 3-dimensional grid of values, which maps the input color values to the corresponding output color values. By applying such a transformation to an image or video, it is possible to achieve a wide range of color and tonal effects, from subtle color grading to dramatic color transformations. 
LUT has already been a classic and commonly used pixel adjustment tool in ISP system~\cite{karaimer2016software} and image editing software because of its high efficiency for modeling color transforms. 
Recently, deep-learning methods based on LUT are proposed in image enhancement tasks. 
Zeng~\etal~\cite{zeng2020learning} first leverage a lightweight CNN to predict the weights for integrating multiple basis LUTs, and the constructed image-adaptive LUT is utilized to achieve image enhancement.
Wang~\etal~\cite{wang2021real} further propose a learnable spatial-aware LUT which considers the global scene and local spatial information. 
Yang~\etal~\cite{yang2022adaint} realize the importance of the sampling strategy so that they design a non-uniform sampling strategy based on learnable adaptive sampling intervals to replace the sub-optimal uniform sampling strategy. At the same time, Yang~\etal~\cite{yang2022seplut} also try to combine 1D LUTs and 3D LUT to promote each other and achieve a more lightweight 3D LUT with better performance.
To the best of our knowledge, LUT has not been adopted in LLVE tasks. In this paper, we will introduce how LUT is naturely suitable for LLVE and enables real-time applications.

\section{Method}

This section provides an overview of the structural intricacies of FastLLVE, as shown in Figure~\ref{fig:architecture}. Input video frames are first encoded into latent features through a lightweight encoder network. Afterwards, the latent features are parallel fed into two modules, namely LUT Generation Module and Video Transformation Module. Specifically, a video-adaptive LUT is generated through the LUT Generation Module, while an intensity map is generated for video transformation. Then, each pixel is enhanced via the IA-LUT transformation with its RGB values and enhancement intensity as the index. Finally, the transformed video is feed into a denoising module for further enhancement. Sections~\ref{subsec:lut} and~\ref{subsec:transform} will focus on the LUT Generation Module and Video Transformation Module, respectively. More structure details about the feature encoder network and denoising module can be found in appendix.

\subsection{LUT Generation Module}
\label{subsec:lut}

\begin{figure}[t]
	\centering
    \begin{overpic}[width=0.99\linewidth]{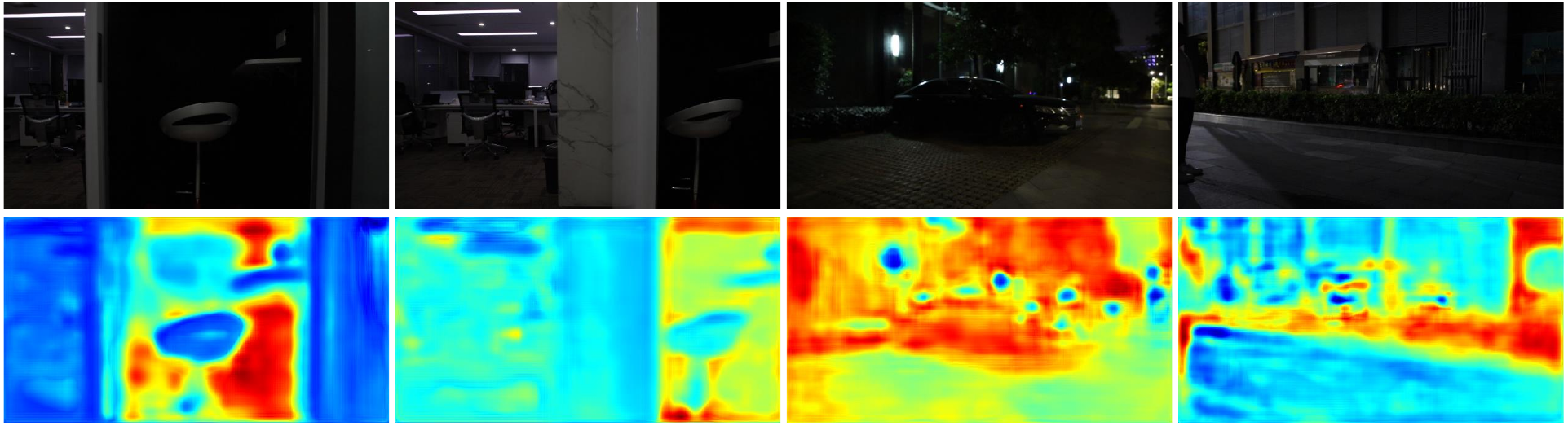}
    \end{overpic}
    \caption{Visualization of the intensity map from extremely low-light videos. The first line consists of input frames and the second line consists of the corresponding intensity maps.}
    \label{fig:intensity}
    \vspace{-4mm}
\end{figure}

\noindent \textit{\textbf{Definition.}}
Although low-light pixels from various areas may appear similar in RGB, they correspond to distinct enhancement intensities during the low-light enhancement process. In Figure~\ref{fig:intensity}, we visualize several intensity maps where even pixels from extremely low-light videos have different enhancement intensities. Traditional 3-dimensional LUTs only save one-to-one mapping relationships for color transformation, which fails on solving the ill-posed problem of low-light pixels with similar color. In order to address this issue, we add a new dimension denoting the enhancement intensity, and the corresponding LUT is denoted as Intensity-Aware LUT. It can store several color spaces for one-to-many mapping relationships and facilitates finer color transformation for LLVE. It is worth noting that only a sampled sparse discrete input space is saved in IA-LUT to avoid introducing massive parameters, which can result in heavy memory burden and great training difficulty. And due to the sparse discrete 4D input space, the LUT transformation of the IA-LUT should be implemented using quadrilinear interpolation.

\begin{figure}[t]
	\centering
	\includegraphics[width=\linewidth]{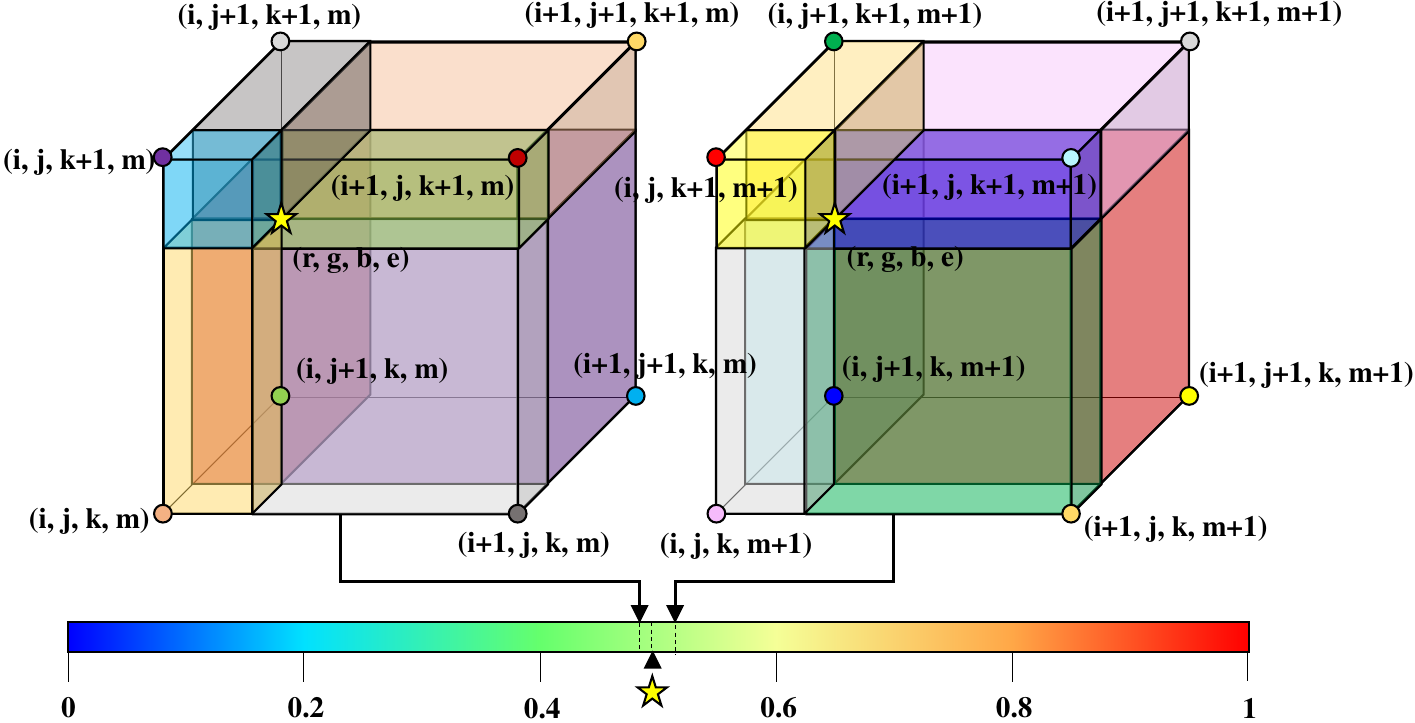}
	\caption{Illustration of the quadrilinear interpolation for one input pixel.}
	\label{fig:interpolation}
    \vspace{-4mm}
\end{figure}

Let $\mathcal{V}: [0, 1]^4 \rightarrow [0, 1]^3$ be a function defined by the IA-LUT, we have 
\begin{equation}
\mathcal{V}(r, g, b, e) = [r', g', b']
\end{equation}
where $r,g,b,e$ indicate the input red, green, blue colors and enhancement intensity, and $r',g',b'$ are the mapped color values.
Let $L$ be the number of grid points in each dimension of the IA-LUT, and $\mathbf{C}_{\mathbf{x}} = [r_{i}, g_{j}, b_{k}, e_{m}]$ stands for the index of grid point $\mathbf{x} = [i,j,k,m]$, where $0 \leq i,j,k,m \leq L-1$. For this grid point $\mathbf{x}$, the stored values in IA-LUT for color mapping are represented as $\mathbf{C}'_{\mathbf{x}} = [r_{\mathbf{x}},g_{\mathbf{x}},b_{\mathbf{x}}]$.
If the input indices $[r,g,b,e]$ can not be mapped to any grid point, we will apply quadrilinear interpolation in the nearest unit lattice.
For brevity, we here let
\begin{equation}
    \Omega_{\mathbf{x}} = (r_{i},r_{i+1}) \times (g_{j},g_{j+1}) \times (b_{k},b_{k+1}) \times (e_{m},e_{m+1}) 
  \end{equation}
as the unit lattice at grid point $\mathbf{x} \in \left\lbrack {0,1,\dots, L - 1} \right\rbrack^{4}$, where we have 
\begin{equation}
    r_{i+1} > r_{i}, g_{j+1} > g_{j}, b_{k+1} > b_{k} \text{ and } e_{m+1} > e_{m}.
\end{equation}
Then, the quadrilinear interpolation process $\mathcal{I}_{\Omega_{\mathbf{x}}}$ in the unit lattice $\Omega_{\mathbf{x}}$ is formulated as:
\begin{equation}
    \mathcal{I}_{\Omega_{\textbf{x}}}(r,g,b,e) = \left[\text{ }\sum\limits_{n = 1}^{2^4}{O_{\mathbf{x}}^n \cdot r_\mathbf{x}^n},\text{ } \sum\limits_{n = 1}^{2^4}{O_{\mathbf{x}}^n \cdot g_\mathbf{x}^n},\text{ } \sum\limits_{n = 1}^{2^4}{O_{\mathbf{x}}^n \cdot b_\mathbf{x}^n}\text{ }\right]\text{ } ,
\end{equation}
where coefficients $O_{\mathbf{x}}^n, n\in [1,2,\dots,16]$ indicates the offsets of the input index to the nearest $2^4$ sampling grids of lattice $\Omega_\mathbf{x}$.
In conclusion, the IA-LUT $\mathcal{V}$ can be formulated as:
\begin{equation}
    \mathcal{V}(r, g, b, e) = 
    \begin{cases}
    \mathbf{C}'_{\mathbf{x}}, &\text{if } [r,g,b,e] = [r_i, g_j, b_k, e_m],\\
    \mathcal{I}_{\Omega_{\mathbf{x}}}(r,g,b,e), &\text{if } [r,g,b,e] \in \Omega_{\mathbf{x}},\\
    [0,0,0], &\text{otherwise.}
    \end{cases}
\end{equation}
In Figure~\ref{fig:interpolation}, we illustrate the quadrilinear interpolation process, and the detailed formulation of coefficients can be found in appendix.

\noindent \textit{\textbf{Generation.}}
In order to automatically generate video-adaptive IA-LUT, as shown in Figure~\ref{fig:architecture}, we learn $T$ learnable basis IA-LUTs and fuse them based on $T$ video-dependent weights, where $T$ is the number of basis LUTs. Compared with directly generating all elements of the video-adaptive IA-LUT via CNN, fusing several basis LUTs is more efficient and computationally inexpensive. More specifically, suppose the low-light video $Y \in \mathbb{R}^{N \times H \times W \times 3}$ with $N$ frames of resolution $W\times H$ is taken as the input, at the beginning, a lightweight encoder with five 3D convolution layers, each with a $3\times 3 \times 3$ kernel size, is used to capture the coarse understanding and some global attributes of the input video $Y$. The output of the encoder is resized to a compact feature vector $\gamma \in \mathbb{R}^{P \times Q}$, which serves as a guide to construct video content-dependent LUT parameters. The size of the feature vector $\gamma$ is due to the two hyper-parameters $P$ and $Q$, which denote the number of pixels and the number of channels before the resizing, respectively. In this paper, we set $P$ to $16$ and $Q$ to $64$ according to the structure of the encoder which can be found in the appendix. After the shared encoder, the weight predictor based on the fully-connected layer maps the compact feature vector $\gamma$ into $T$ dynamic video-dependent weights, which can be formulated as:
\begin{equation}
    \left. h_{0}:\mathbb{R}^{P \times Q}\rightarrow \mathbb{R}^{T} \right.,
\end{equation}
where $h_{0}$ denotes the mapping from the feature vector $\gamma$ to the video-dependent weights for fusion. Subsequently, another fully-connected layer is employed to map the video-dependent weights to all elements of the video-adaptive IA-LUT. The learnable parameters of this layer are encoded basis IA-LUTs. We refer to this mapping as $h_{1}$ and describe it as:
\begin{equation}
    \left. h_{1}:\mathbb{R}^{T}\rightarrow \mathbb{R}^{L \times L \times L \times L \times 3} \right.,
\end{equation}
where $L \times L \times L \times L \times 3 = 3\times L^4$ is the total number of elements of the generated video-adaptive IA-LUT, and the number $3$ means that the IA-LUT stores the mapped red, green and blue color values, respectively. The elements of the basis IA-LUTs can be updated during the end-to-end training since they serve as the parameters of the fully-connected layer, which makes the basis LUTs learnable.

In general, besides the shared encoder, two fully-connected layers achieve the main mapping $h$ from the feature vector $\gamma$ to the generated video-adaptive IA-LUT, as shown below:
\begin{equation}
    h:\gamma \overset{h_{0}}{\rightarrow}w \in \mathbb{R}^{T}\overset{h_{1}}{\rightarrow}\mathbf{C}' \in \mathbb{R}^{L \times L \times L \times L \times 3},
\end{equation}
where $\mathbf{C}' \in \mathbb{R}^{L \times L \times L \times L \times 3}$ denotes all the stored elements $\mathbf{C}'_{\mathbf{x}}$ of the target video-adaptive IA-LUT, and $w \in \mathbb{R}^{T}$ represents the video-dependent weights obtained through the mapping $h_{0}$. As shown above, the mapping $h$ is actually a cascade of the mapping $h_{0}$ and $h_{1}$. It's worth emphasizing that dividing the main mapping $h$ into two parts, each realized through a fully-connected layer, is crucial to reduce the number of parameters, similar to the sampled input space of LUT. Using only one fully-connected layer to directly map the compact feature vector $\gamma$ to the generated LUT would lead to a significantly larger number of parameters, specifically $P \times Q \times 3\times L^4$, compared to $T \times \left( {P \times Q + 3\times L^4} \right)$. Therefore, dividing the mapping $h$ by rank factorization and implementing it with two fully-connected layers can reduce the parameters, making the transformation easier to learn and optimize.

\begin{table*}[t]
  \centering
  \caption{Quantitative comparisons on SDSD and SMID test datasets. Top two numbers of each column are with the best in \textcolor[rgb]{ 1,  0,  0}{red} and the second best in \normalfont{\textcolor[rgb]{ 0,  0,  1}{blue}}. \textbf{"FastLLVE+\emph{dn}" denotes our method with a simple denoising module.}}
  \setlength{\tabcolsep}{2.8mm}{
    \begin{tabular}{ccccccccccc}
    \toprule
    \multirow{2}{*}{Format} & \multirow{2}{*}{Method} & \multicolumn{4}{c}{SDSD}      & \multicolumn{4}{c}{SMID}      & \multirow{2}{*}{Runtime (s)} \\
\cmidrule{3-10}          &       & PSNR  & SSIM  & AB~(Var)↓ & MABD↓ & PSNR  & SSIM  & AB~(Var)↓ & MABD↓ &  \\
    \midrule
    \multirow{2}{*}{Image} & MBLLEN \cite{lv2018mbllen} & 21.79 & 0.65  & \textbackslash{} & \textbackslash{} & 22.67 & 0.68  & \textbackslash{} & \textbackslash{} & 0.336 \\
          & Cheby \cite{pan2022chebylighter} & 23.65 & \textcolor[rgb]{ 0,  0,  1}{0.81} & 0.079 & 0.297 & 25.24 & 0.76  & 1.486 & 1.891 & 0.175 \\
    \midrule
    \multirow{5}{*}{Video} & SALVE \cite{azizi2022salve} & 18.03 & 0.69  & 0.125 & 0.246 & 16.73 & 0.60  & 1.984 & 3.501 & 0.182 \\
          & SMOID \cite{jiang2019learning} & 23.45 & 0.69  & 0.397 & 0.749 & 23.64 & 0.71  & 1.455 & 1.736 & 0.190 \\
          & SMID \cite{chen2019seeing} & 24.09 & 0.69  & 0.784 & 1.592 & 24.78 & 0.72  & \textcolor[rgb]{ 0,  0,  1}{0.405} & 0.794 & 0.125 \\
          & SDSDNet \cite{wang2021seeing}  & 24.92 & 0.73  & 0.181 & 0.193 & 26.03 & 0.75  & 0.737 & 0.944 & 0.407 \\
          & StableLLVE \cite{zhang2021learning} & 23.09 & \textcolor[rgb]{ 0,  0,  1}{0.81} & 1.366 & 2.814 & 26.22 & \textcolor[rgb]{ 0,  0,  1}{0.78}  & 0.745 & 0.897 & \textcolor[rgb]{ 0,  0,  1}{0.038} \\
    \midrule
    \multirow{2}{*}{Video} & FastLLVE  & \textcolor[rgb]{ 0,  0,  1}{27.06} & 0.78  & \textcolor[rgb]{ 0,  0,  1}{0.038} & \textcolor[rgb]{ 0,  0,  1}{0.091} & \textcolor[rgb]{ 0,  0,  1}{26.45} & 0.75 & 0.476 & \textcolor[rgb]{ 0,  0,  1}{0.748} & \textcolor[rgb]{ 1,  0,  0}{\textbf{0.013}} \\
          & FastLLVE+\emph{dn} & \textcolor[rgb]{ 1,  0,  0}{\textbf{27.55}} & \textcolor[rgb]{ 1,  0,  0}{\textbf{0.86}} & \textcolor[rgb]{ 1,  0,  0}{\textbf{0.033}} & \textcolor[rgb]{ 1,  0,  0}{\textbf{0.040}} & \textcolor[rgb]{ 1,  0,  0}{\textbf{27.62}} & \textcolor[rgb]{ 1,  0,  0}{\textbf{0.80}} & \textcolor[rgb]{ 1,  0,  0}{\textbf{0.065}} & \textcolor[rgb]{ 1,  0,  0}{\textbf{0.050}} & 0.080 \\
    \bottomrule
    \end{tabular}}%
  \label{tab:tabel2}%
  \vspace{-4mm}
\end{table*}%

\subsection{Video Transformation}
\label{subsec:transform}
As shown in Figure~\ref{fig:architecture}, we first estimate the intensity map, then perform look-up according to the RGB video and corresponding intensity map. In order to construct this map, a lightweight decoder with five deconvolution layers~\cite{zeiler2014visualizing} of size $3 \times 3 \times 3$  is adopted to utilize the latent features from the encoder, resulting in an intensity map $I \in \mathbb{R}^{N \times H \times W \times 1}$. By concatenating the intensity map and the input video, we can perform look-up and interpolation as introduced in Section~\ref{subsec:lut}. We recursively apply the transformation on each frame of a video sequence, resulting in the video output with stable and consistent brightness.

In addition, as the LUT transformation is applied to each pixel independently and quadrilinear interpolation can be parallel processed, we implement the IA-LUT transformation via CUDA to accelerate the transformation and achieve the convenient end-to-end training. Specifically, we merge the lookup and interpolation operations into a single CUDA kernel to maximize the parallelism. Following Adaint~\cite{yang2022adaint}, we also adopt binary search algorithm during lookup operation, because the logarithmic time complexity can make computational cost negligible, unless $L$ is set to an unexpected large value. It is important to emphasize that the pixel-wise transformation, which is indexed only by the red, green, blue colors and enhancement intensity of each input pixel, is the key to the IA-LUT naturally maintaining the inter-frame brightness consistency.

Although the FastLLVE framework can naturally maintain inter-frame brightness consistency, and achieves the great performance at the same time, it should be pointed out that LUT is susceptible to noise. In the real world, images and videos captured in low-light conditions inevitably contain noises. Therefore, we sacrifice some efficiency to add a simple denosing module as the post-processing, refining the enhanced normal-light video for better performance. Specifically, in this paper, we choose to follow the practice in~\cite{chen2019real} to design the additional denoising module. However, it is worth noting that almost all existing denoising methods, such as~\cite{mao2016image, zhang2017beyond, ren2019dn, tassano2020fastdvdnet}, can be alternatives as the post-processing. 
\section{Experiments}

\subsection{Implementation Details}
We implement our method based on PyTorch~\cite{paszke2019pytorch} and train the framework on a NVIDIA GeForce RTX 3090 GPU. The standard Adam optimizer~\cite{kingma2014adam} is adopted to train the entire network, with the batch size set to 8. The initial learning rate is set to $4 \times 10^{- 4}$ and gradually decayed according to the scheme of Cosine annealing~\cite{gotmare2018closer} with restart set to $10^{-7}$.

Regarding the loss function, since previous LUT-based methods~\cite{zeng2020learning, yang2022adaint, yang2022seplut} have proven the effectiveness of the smooth regularization and monotonicity regularization, we add 4D smooth regularization and 4D monotonicity regularization adapted to the IA-LUT into the loss function. If we add the additional denoising module, a pairwise loss between the denoised result and ground truth will be included in the loss function. As a result, the total loss function is defined as:
\begin{equation}
    l_{total} = 
    \begin{cases}
    l_{r0} + l_{r1} + \alpha_{s}l_{s} + \alpha_{m}l_{m}, &\text{if denoising},\\
    l_{r0} + \alpha_{s}l_{s} + \alpha_{m}l_{m}, &\text{otherwise},
    \end{cases}
\end{equation}
 where $l_{r0}$ denotes the reconstruction loss between the transformed normal-light video and ground truth, and $l_{r1}$ denotes the denoising loss between the final denoised result and ground truth. Both of them use Charbonnier Loss~\cite{lai2017deep, lai2018fast}. The 4D smooth regularization loss $l_{s}$ consists of two parts which correspond to the video-adaptive IA-LUT and the video-dependent weights, respectively. It prevents artifacts caused by extreme color changes in LUT, while the 4D monotonicity regularization loss $l_{m}$ preserves the robustness during the enhancement process. The detailed formulations of $l_{s}$ and $l_{m}$ can be found in appendix.

As for hyper-parameters, in the loss function, we set $\alpha_{s}$ and $\alpha_{m}$ to 0.0001 and 10, respectively. In terms of $L$ and $T$, although higher values contribute to the precision of color transformation modeled by LUT, they can significantly increase the parameters of all LUTs used in the entire framework. Therefore, we follow the most widely-used setting~\cite{zeng2020learning, yang2022adaint, yang2022seplut} of the two numbers, which is proposed to balance the accuracy and the size of parameters. Thus, the number of sampling grid points on each dimension is set to 33 and the number of basis IA-LUTs is set to 3. 


\subsection{Experiment Setup}
We present a comprehensive comparison of our proposed method with eight SOTA low-light enhancement methods, including both image-based and video-based approaches. The image-based methods we evaluate are MBLLEN~\cite{lv2018mbllen}, Cheby~\cite{pan2022chebylighter} and LEDNet~\cite{zhou2022lednet}, while SMID~\cite{chen2019seeing}, SMOID~\cite{jiang2019learning}, StableLLVE~\cite{zhang2021learning}, SALVE~\cite{azizi2022salve} and SDSDNet~\cite{wang2021seeing} are completely video-based methods. We use their released codes and follow the same training strategies to train these networks on two real-world low-light video datasets, namely SDSD~\cite{wang2021seeing} dataset and SMID~\cite{chen2019seeing} dataset. The SDSD dataset is split into the SDSD training and test datasets with $23,542$ and $750$ video frames respectively, while the SMID training and test datasets contain $18,278$ and $1,470$ video frames. Then, we evaluate the performance of FastLLVE and the compared methods on the SDSD and SMID test datasets, to demonstrate the superiority of our method. It's worth noting that the SMID dataset used in our experiments has been pre-processed to transform the RAW data into sRGB data via its own pre-processing method proposed in SMID.

\begin{figure}[t]
	\centering
	\includegraphics[width=\linewidth]{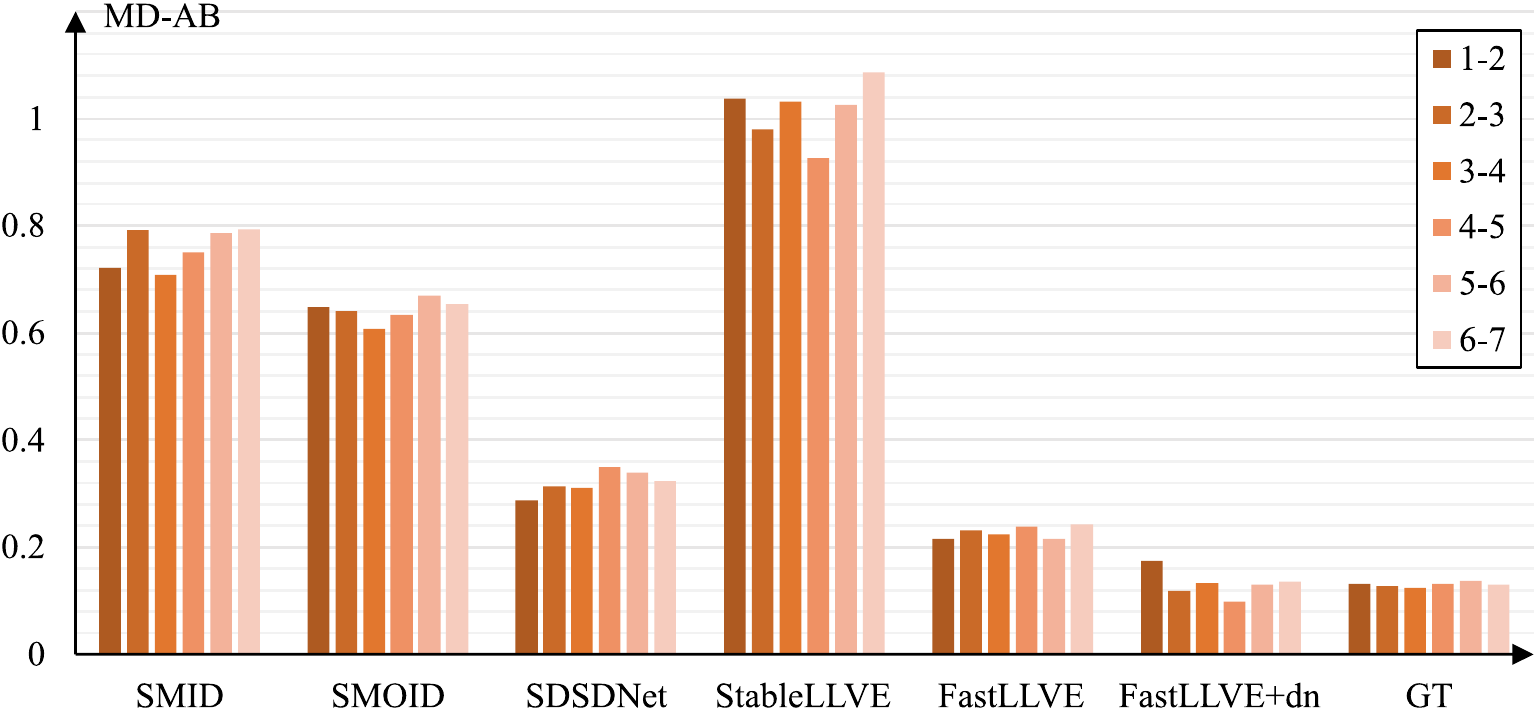}
	\caption{The Mean Differences of Average Brightness (MD-AB) between adjacent frames of SDSD test dataset. [Key: 1-2: difference between the 1st frame and the 2nd frame, 2-3 to 6-7 are indicated similarly]}
	\label{fig:brightness}
	\vspace{-2mm}
\end{figure}
We use the common evaluation metrics of Peak Signal-to-Noise Ratio (PSNR) and Structural Similarity (SSIM) to assess the quality of the enhanced videos. In addition, drawing on previous works~\cite{lv2018mbllen, jiang2019learning, zhang2021learning}, we consider the variance of Average Brightness (AB (Var)) and Mean Absolute Brightness Difference (MABD) to evaluate the ability to maintain inter-frame brightness consistency, where lower values stand for better consistency. Furthermore, we also record the average processing time of a 1,080p low-light frame on a NVIDIA GeForce RTX 3090 GPU for each method.

\subsection{Comparisons of Brightness Consistency}
As presented in Table~\ref{tab:tabel2}, the AB (Var) and MABD are employed to assess the maintenance of inter-frame brightness consistency. FastLLVE+\emph{dn} outperforms the compared methods on the two test datasets, including the SDSDNet that is the current SOTA in terms of brightness consistency. \textit{It is worth noting that SDSDNet also contains a denoising module based on the Retinex theory}~\cite{land1977retinex}. Moreover, we also compute the Mean Differences of Average Brightness (MD-AB) between adjacent frames of SDSD test dataset shown in Figure~\ref{fig:brightness}. It is evident that the FastLLVE+\emph{dn} achieves the best performance in brightness consistency and behaves most similar to ground truth.

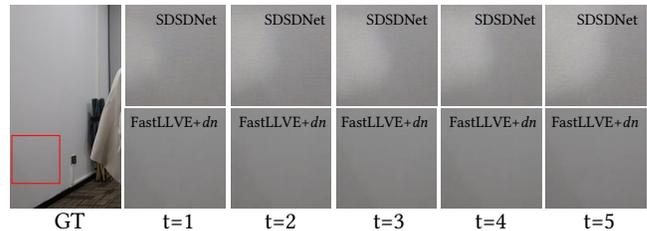
\begin{figure}[t]
	\centering
    \begin{overpic}[width=\linewidth]{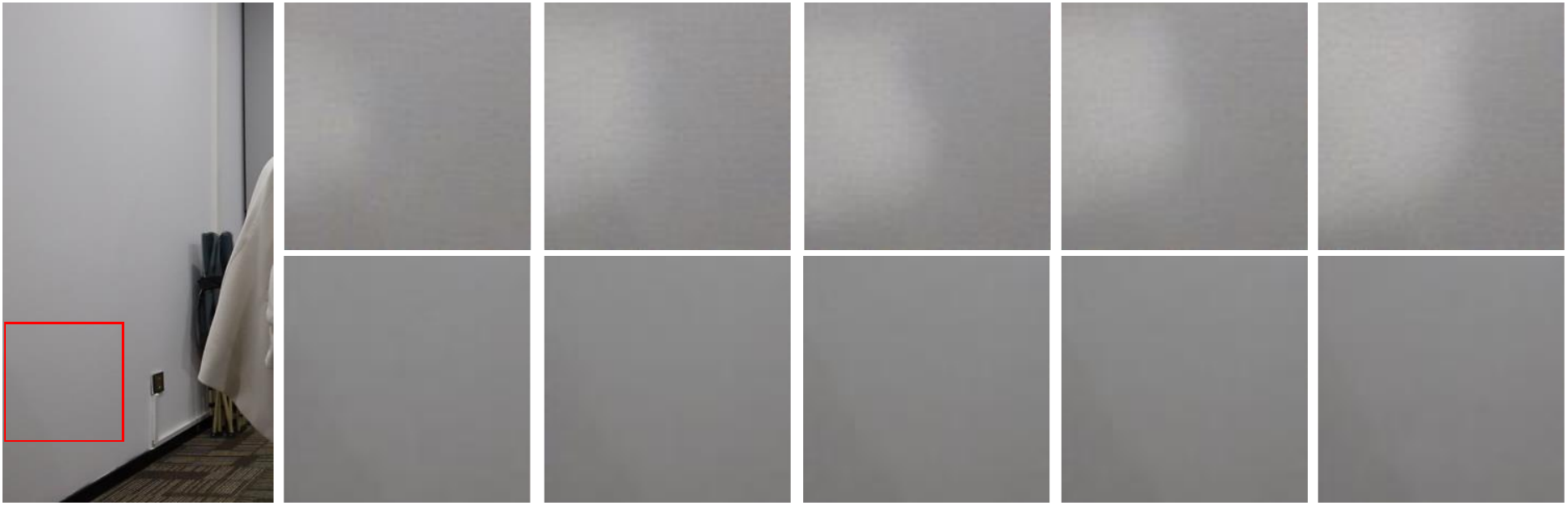}
        \put(23,29){\scriptsize SDSDNet}
        \put(40,29){\scriptsize SDSDNet}
        \put(56,29){\scriptsize SDSDNet}
        \put(73,29){\scriptsize SDSDNet}
        \put(89,29){\scriptsize SDSDNet}
        \put(19,13){\scriptsize FastLLVE+\emph{dn}}
        \put(36,13){\scriptsize FastLLVE+\emph{dn}}
        \put(52,13){\scriptsize FastLLVE+\emph{dn}}
        \put(69,13){\scriptsize FastLLVE+\emph{dn}}
        \put(85,13){\scriptsize FastLLVE+\emph{dn}}
        \put(7,-3){GT}
        \put(24,-3){t=1}
        \put(40,-3){t=2}
        \put(57,-3){t=3}
        \put(73,-3){t=4}
        \put(90,-3){t=5}
    \end{overpic}
    \vspace{1mm}
    \caption{The local inter-frame brightness inconsistency on SDSD test dataset.}
    \label{fig:inconsistency}
    \vspace{-4mm}
\end{figure}
In addition to quantitative comparisons, we also provide the qualitative comparisons in Figure~\ref{fig:inconsistency}. SDSDNet generates an incorrect light spot that varies among adjacent frames, which manifests its limitation when dealing with local brightness inconsistencies. In comparison, FastLLVE+\emph{dn} has better ability to suppress the inter-frame brightness inconsistency in local areas, benefiting from the elaborete design of IA-LUT.

\subsection{Comparisons of Enhanced Performance}
In Table~\ref{tab:tabel2}, we present the performance comparisons in terms of PSNR and SSIM on both SDSD and SMID test datasets.
Compared with the StableLLVE, our FastLLVE achieves almost $3\times$ inference speed and outperforms it in terms of PSNR. Meanwhile, the FastLLVE+\emph{dn} achieves the SOTA performance in PSNR and SSIM on both datasets, and outperforms the StableLLVE by a large margin. Notably, the generally higher values of PSNR on SDSD test dataset indicate the difficulty of color restoration from the color space with an extremely low dynamic range since most videos in SDSD test dataset are taken in extremely low-light scenarios.

\begin{figure*}[t]
	\centering
	\begin{overpic}[width=\linewidth]{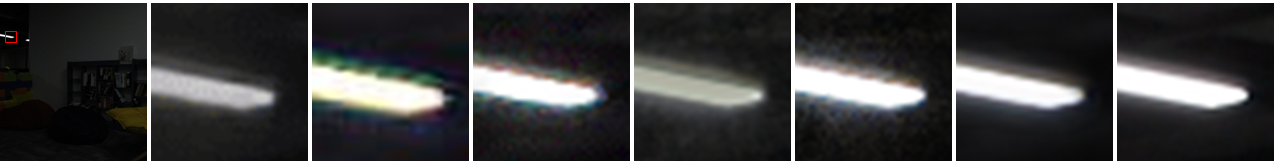}
		\put(4,-2){Input}
        \put(16,-2){SMOID}
        \put(29,-2){SMID}
        \put(40,-2){SDSDNet}
        \put(52,-2){StableLLVE}
        \put(65,-2){FastLLVE}
        \put(77,-2){FastLLVE+\emph{dn}}
        \put(93,-2){GT}
	\end{overpic}
    \vspace{-1mm}
    \caption{Qualitative comparisons on SDSD test dataset. Our method achieves correct enhancement and the best performance.}
    \label{fig:demo2}
    \vspace{-2mm}
\end{figure*}

\begin{figure*}[t]
	\centering
    \begin{overpic}[width=\linewidth]{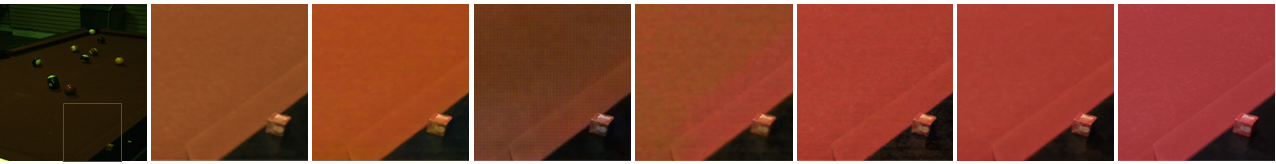}
        \put(4,-2){Input}
        \put(16,-2){SMOID}
        \put(29,-2){SMID}
        \put(40,-2){SDSDNet}
        \put(52,-2){StableLLVE}
        \put(65,-2){FastLLVE}
        \put(77,-2){FastLLVE+\emph{dn}}
        \put(93,-2){GT}
    \end{overpic}
    \vspace{-1mm}
    \caption{Qualitative comparisons on SMID test dataset. Compared to other methods, only our method restores color correctly.}
    \label{fig:demo3}
    \vspace{-4mm}
\end{figure*}
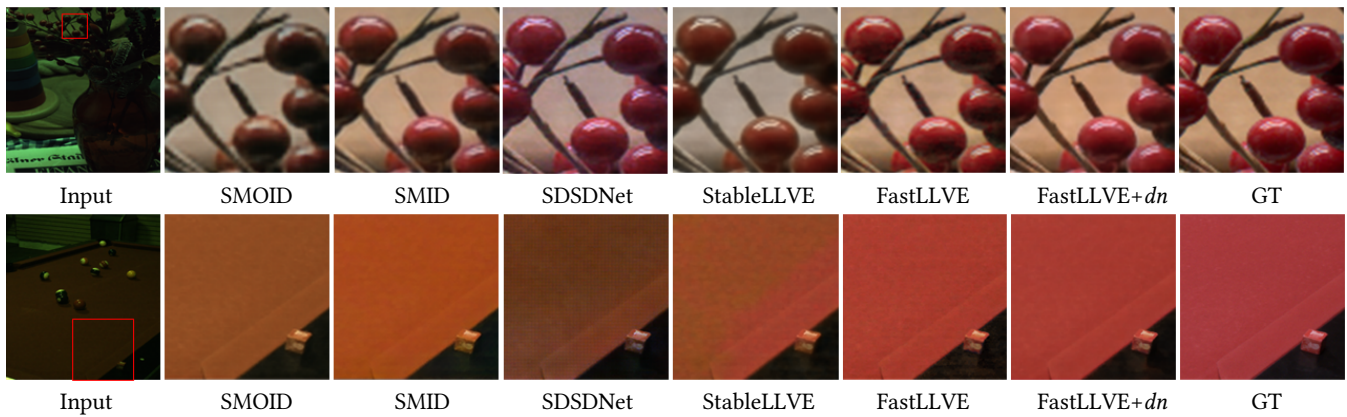

We further conduct qualitative comparisons on the two datasets in Figure~\ref{fig:demo2}-\ref{fig:demo3}. In Figure~\ref{fig:demo2}, we present the results of extremely low-light video in SDSD test dataset. Except for SMOID and SDSDNet, previous methods suffer from incorrect color restoration evidently. SMOID produces blurry results and lack of texture details. SDSDNet produces certain degree of noise owing to the deviation of noise map estimation. Besides, the darker area and the colorful border of the bright area appear in the results of other methods, which indicates the incorrect enhancement in brightness. Compared with these methods, our FastLLVE+\emph{dn} solves the difficulties of color restoration, enhancement in brightness, as well as noise suppression, achieving visually great performance. In Figure~\ref{fig:demo3}, it is challenging to restore low-light videos with severe color biases. For instance, the enhanced low-light videos from SDSDNet present much darker than the ground truth, and the results of StableLLVE suffer from color distortions. The reason behind might be the inaccurate estimation of color transformation. In comparison, both FastLLVE and FastLLVE+\emph{dn} can enhance the low-light videos well with a similar color space to ground truth, even though the color biases exist.

\begin{figure*}[t]
	\centering
        \begin{overpic}[width=\linewidth]{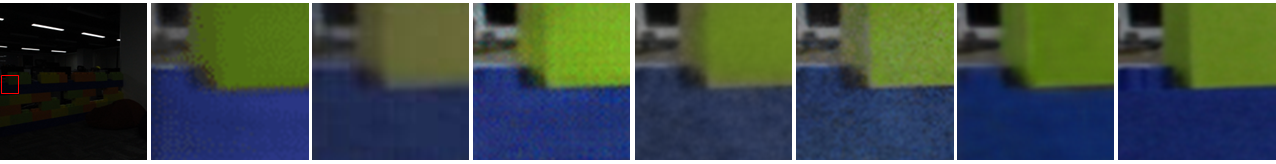}
		\put(4,-2){Input}
        \put(16,-2){SMOID}
        \put(29,-2){SMID}
        \put(40,-2){SDSDNet}
        \put(52,-2){StableLLVE}
        \put(65,-2){FastLLVE}
        \put(77,-2){FastLLVE+\emph{dn}}
        \put(93,-2){GT}
	\end{overpic}
    \vspace{-2mm}
\end{figure*}
\begin{figure*}[t]
	\centering
    \begin{overpic}[width=\linewidth]{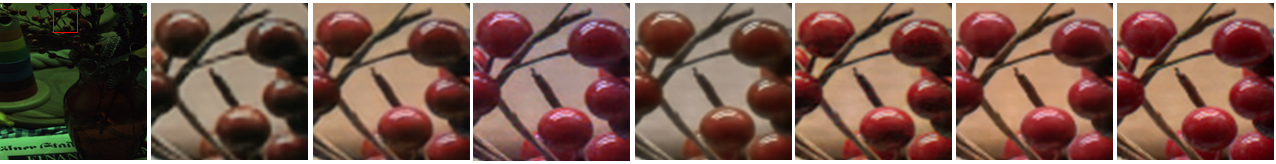}
        \put(4,-2){Input}
        \put(16,-2){SMOID}
        \put(29,-2){SMID}
        \put(40,-2){SDSDNet}
        \put(52,-2){StableLLVE}
        \put(65,-2){FastLLVE}
        \put(77,-2){FastLLVE+\emph{dn}}
        \put(93,-2){GT}
    \end{overpic}
    \vspace{-2mm}
\end{figure*}
\subsection{Ablation Study}
\begin{figure*}[t]
	\centering
    \begin{minipage}[h]{0.24\linewidth}
		\centering
		\includegraphics[width=\linewidth]{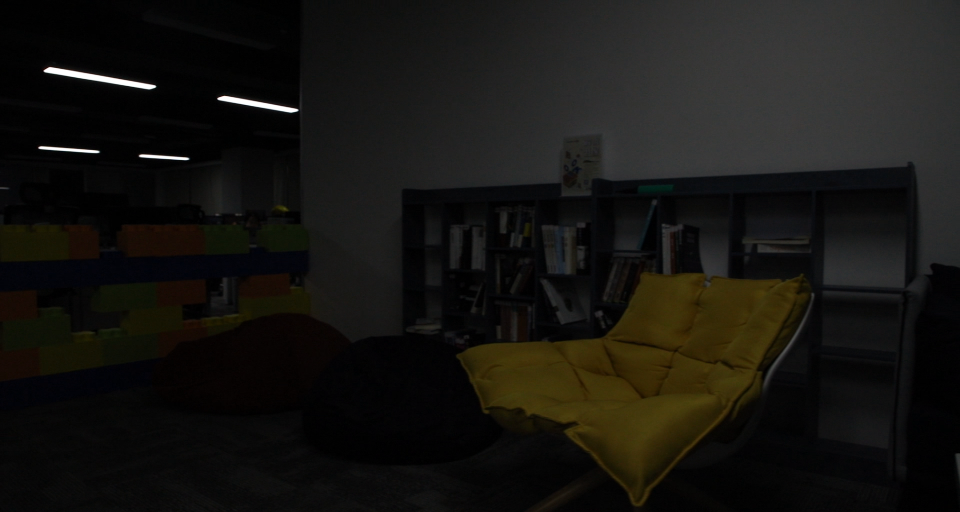}
            \small{(a) Input}
	\end{minipage}
	\begin{minipage}[h]{0.24\linewidth}
		\centering
		\includegraphics[width=\linewidth]{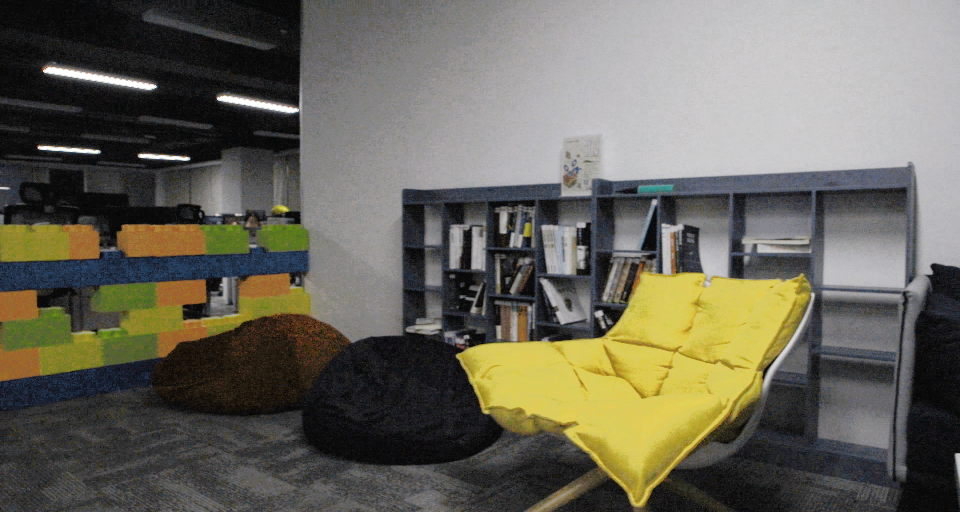}
            \small{(b) 3D LUT}
	\end{minipage}
	\begin{minipage}[h]{0.24\linewidth}
		\centering
		\includegraphics[width=\linewidth]{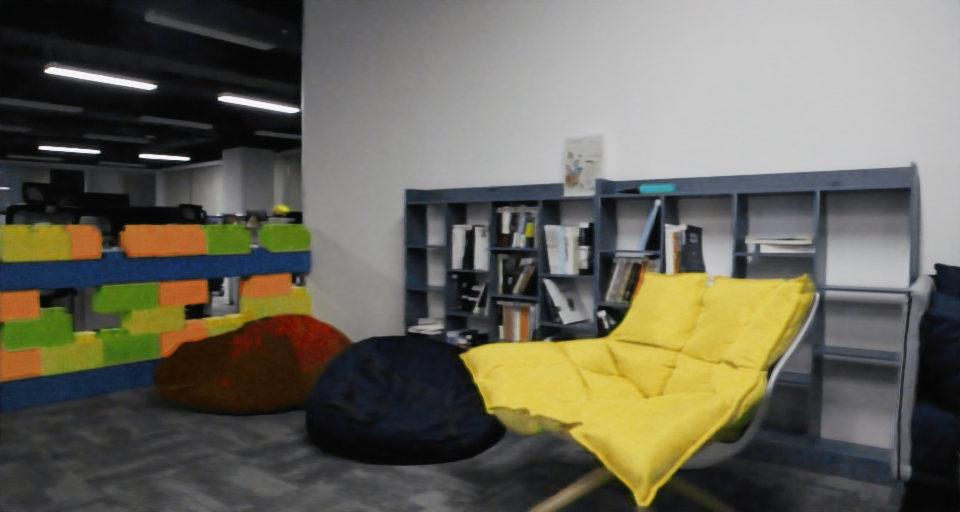}
            \small{(c) IA-LUT}
	\end{minipage}
	\begin{minipage}[h]{0.24\linewidth}
		\centering
		\includegraphics[width=\linewidth]{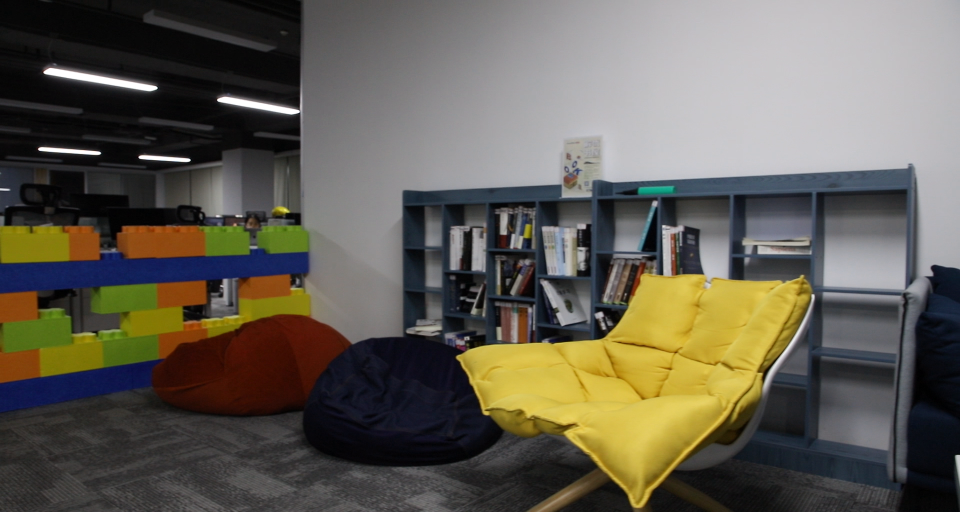}
            \small{(d) GT}
	\end{minipage}
	\begin{minipage}[h]{0.24\linewidth}
		\centering
		\includegraphics[width=\linewidth]{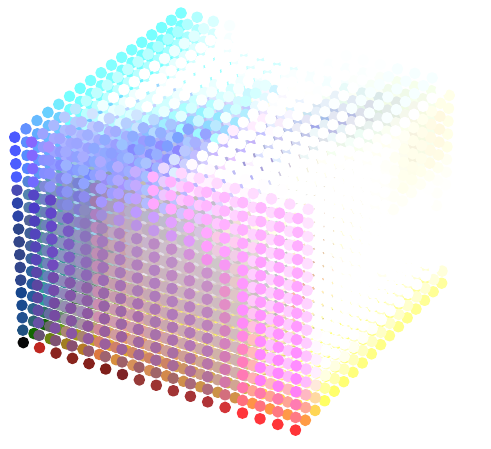}
            \small{(e) 3D LUT}
	\end{minipage}
	\begin{minipage}[h]{0.24\linewidth}
		\centering
		\includegraphics[width=\linewidth]{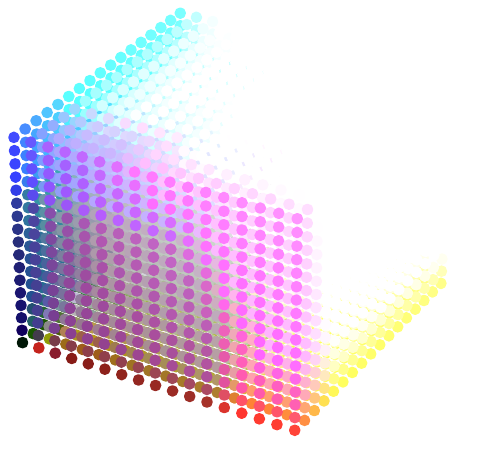}
            \small{(f) IA-LUT ($e = 0$)}
	\end{minipage}
	\begin{minipage}[h]{0.24\linewidth}
		\centering
		\includegraphics[width=\linewidth]{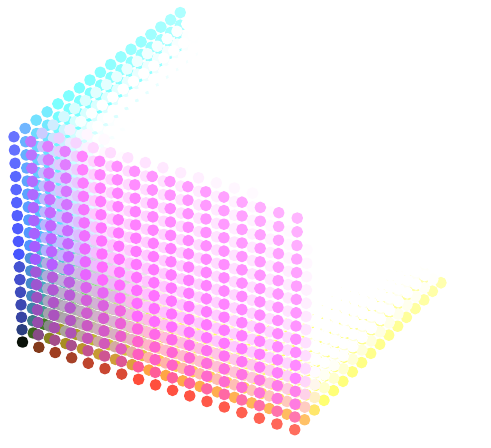}
            \small{(g) IA-LUT ($e = 0.5$)}
	\end{minipage}
	\begin{minipage}[h]{0.24\linewidth}
		\centering
		\includegraphics[width=\linewidth]{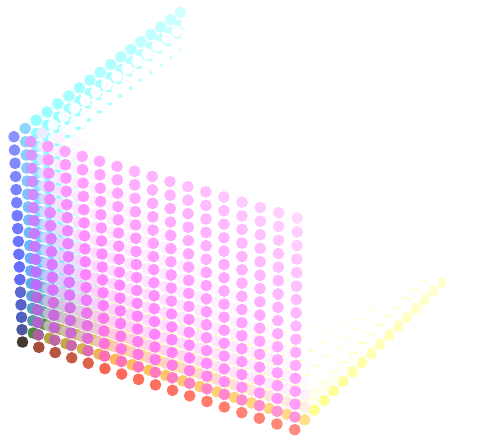}
            \small{(h) IA-LUT ($e = 1$)}
	\end{minipage}
	\caption{Visualization of the common 3D LUT and the remainder of IA-LUT while the enhancement intensities $e$ are fixed.}
	\label{fig:LUT}
    \vspace{-4mm}
\end{figure*}

In order to evaluate the effectiveness of the IA-LUT and explore better use of the denoising module, we perform the ablation study and show the results in Figure~\ref{fig:LUT} and Table~\ref{tab:table3}.

\noindent\textbf{Structure of LUT:} We change IA-LUT into the common 3D LUT and remove the decoder, so as to construct the low-light video enhancement network based on 3D LUT in comparison to the FastLLVE with IA-LUT. As shown in Table~\ref{tab:table3}, FastLLVE completely outperforms the 3D LUT-based network, which demonstrates that the additional dimension of enhancement intensity greatly improves the ability of common 3D LUT to model color transformation under low-light conditions. 

In addition to quantitative comparisons, we also visualize the common 3D LUT and our IA-LUT in Figure~\ref{fig:LUT}. Since 4-dimension IA-LUT is difficult to illustrate succinctly, we select out three representative enhancement intensities $e$ and fix them to draw the 3D remainder in IA-LUT. From the visualization, it can be seen that the remainder of IA-LUT models the color transformation more smoothly and correctly than 3D LUT. More importantly, with $e$ increasing, the 3D remainder of IA-LUT that stores the mapping relationships of colors is inclined to be brighter. This observation validates the influence of enhancement intensities on determining the relatively optimal color transformation, which is in line with our overall design.

\noindent\textbf{Denoising:} As we pointed out, LUT is susceptible to noise. However, the real-world videos captured in low-light conditions are often degraded by noises. To this end, it is expected to improve the network performance by adding a denoising module to suppress the noises. In the comparisons between FastLLVE and FastLLVE+\emph{dn}, a simple denoising module is able to complement the LUT in performance with the evident improvement on all metrics. Note that the denoising module also benefits the evaluation of brightness consistency. One possible reason is that the AB (Var) and MABD are sensitive to noise as well.

\begin{table*}[t]
  \centering
  \caption{The results of ablation study. The best is in \textcolor[rgb]{ 1,  0,  0}{red}. [Key: 3D LUT: baseline based on 3D LUT, FastLLVE: our framework without denoising module, \emph{dn}+FastLLVE: our framework with denoising module as the pre-processing, FastLLVE+\emph{dn}: our framework with denoising module as the post-processing]}
  \setlength{\tabcolsep}{3.5mm}{
    \begin{tabular}{cccccccccc}
    \toprule
    \multirow{2}{*}{Method} & \multicolumn{4}{c}{SDSD}      & \multicolumn{4}{c}{SMID}      & \multirow{2}{*}{Runtime(s)} \\
\cmidrule{2-9}          & PSNR  & SSIM  & AB (Var)↓ & MABD↓ & PSNR  & SSIM  & AB (Var)↓ & MABD↓ &  \\
    \midrule
    3D LUT & 22.76 & 0.69  & 0.107 & 0.265 & 25.08 & 0.73  & 1.024 & 1.227 & \textcolor[rgb]{ 1,  0,  0}{\textbf{0.007}} \\
    FastLLVE & 27.06 & 0.78  & 0.038 & 0.091 & 26.45 & 0.75  & 0.476 & 0.748 & 0.013 \\
    \emph{dn}+FastLLVE & 24.02 & 0.81  & 0.146 & 0.368 & 23.85 & 0.72  & 1.657 & 2.872 & 0.080 \\
    FastLLVE+\emph{dn} & \textcolor[rgb]{ 1,  0,  0}{\textbf{27.55}} & \textcolor[rgb]{ 1,  0,  0}{\textbf{0.86}} & \textcolor[rgb]{ 1,  0,  0}{\textbf{0.033}} & \textcolor[rgb]{ 1,  0,  0}{\textbf{0.040}} & \textcolor[rgb]{ 1,  0,  0}{\textbf{27.62}} & \textcolor[rgb]{ 1,  0,  0}{\textbf{0.80}} & \textcolor[rgb]{ 1,  0,  0}{\textbf{0.065}} & \textcolor[rgb]{ 1,  0,  0}{\textbf{0.050}} & 0.080 \\
    \bottomrule
    \end{tabular}}%
  \label{tab:table3}%
  \vspace{-2mm}
\end{table*}%

\noindent\textbf{Locations of denoising module:} Except for setting a denoising module as the post-processing, it is also feasible to construct a denoising module at the begining of the whole framework as the pre-processing. Neglect to the possibility of enhancing noise, the pre-processing denoising module seems to be more reasonable. However, it is worth noting that there is no feasible way to collect clean low-light videos in real world, which leads to the lack of ground truths for the supervised training of the denoising module as the pre-processing. As a consequence, the denoising module without supervision can affect the video-adaptive IA-LUT to learn incorrect mapping through the end-to-end training of the entire framework. In addition, even though the training principles~\cite{lehtinen2018noise2noise, moran2020noisier2noise} are specifically designed for supervised training with unpaired noisy data, they may still not be able to completely solve this problem since it is too difficult to find a reliable noise model that enables the simulation in various low-light conditions.

For comparison, we follow the training principle Noiser2Noise \cite{moran2020noisier2noise} to self-supervise the denoising module as the pre-processing, with the low-light noise model from~\cite{lv2021attention} simulating the noise in low-light conditions. As shown in Table~\ref{tab:table3}, FastLLVE+\emph{dn} performs much better than \emph{dn}+FastLLVE on all metrics, which supports the denoising module as the post-processing. Otherwise, due to the unreliable low-light noise model, the training of \emph{dn}+FastLLVE based on Noiser2Noise is unstable and easy to fail as expected. Therefore, we use the denosing module as the post-processing in our method.
\section{Limitations and Broader Impact}
In this paper, we propose a Intensity-Aware LUT for LLVE and validate its advantages of high efficiency and natural maintenance of inter-frame brightness consistency. However, since the LUT-based methods is commonly susceptible to noise, a from-the-shelf denoising module is leveraged to further improve the visual quality of enhanced normal-light videos, affecting the efficiency of the whole framework. Therefore, a novel denoising strategy suitable for LUT is promising, and we leave it as future work.

Real-time LLVE has potential to bring a significant impact in many ways. On the one hand, it can by applied on camera monitor system to improve the public safety by increasing the visibility of critical areas, such as streets, parking lots, and transportation, particularly during night-time hours. On the other hand, it can enhance the quality of visual media produced in low-light conditions, such as documentary footage and home videos. This would improve the overall quality of such visual media, making them more engaging and informative.

\section{Conclusion}
We firstly introduce the LUT in LLVE tasks, and propose a novel LUT-based framework, named FastLLVE, for real-time low-light video enhancement. In order to bring the LUT to LLVE, we design the Intensity-Aware LUT (IA-LUT) with a new dimension of enhancement intensity to solve the one-to-many mapping problem. In terms of the flickering effect in the enhanced video, we point out that IA-LUT can naturally maintain the brightness consistency among video frames. Extensive experiments have validated the superiority of the proposed method as compared to the LLVE SOTAs. We envision that this work could facilitate the development of LLVE in practical applications.

\section*{Acknowledgement}
This work was supported in part by the Shanghai Pujiang Program under Grant 22PJ1406800.

\bibliographystyle{ACM-Reference-Format}
\bibliography{sample-base}

\renewcommand\thesection{\Alph{section}}
\setcounter{section}{0}
\section{Implementation Details}
\label{sec:net}
\textit{\textbf{Network Structure.}}
As illustrated in Table~\ref{tab:table1}, the lightweight encoder is comprised of five convolutional blocks that perform down-sampling on each frame of the input video, thereby reducing the resolution to 1/32. Each convolutional block consists of a 3D convolution layer, a leaky ReLU~\cite{xu2015empirical} activation function, and an instance normalization~\cite{ulyanov2016instance} layer. Besides, the corresponding lightweight decoder consists of five deconvolutional blocks that restore each frame of the encoder output to the original resolution through up-sampling. The deconvolutional block has the same structure as the convolutional block, with the exception that the convolution layer is replaced by a deconvolution~\cite{zeiler2014visualizing} layer. It is worth noting that all instance normalization layers used in the network have learnable parameters, and the last deconvolutional block that directly outputs the intensity map is devoid of the leaky ReLU activation function and instance normalization layer.

In addition, before the two fully-connected layers, which achieve the mapping h from the compact feature vector $\gamma$ to the video-adaptive IA-LUT, the latent features extracted from the encoder are first subjected to an average pooling operation, followed by a dropout operation with the dropout rate set to 0.5. Subsequently, the features are reshaped to obtain the feature vector $\gamma$. In terms of these operations as the pre-processing before weight predictor, the average pooling operation serves to reduce the parameters of the first fully-connected layer, and the dropout operation aims at enhancing training data, instead of just avoiding over-fitting that is the common role of dropout operation.

As for the denoising module, in this paper, we follow the practice in~\cite{chen2019real} to design the additional denoising module. Compared with the original DDFN approach, we decrease the number of feature integration blocks to one, thereby reducing the size of the denoising module. Moreover, in order to effectively process the enhanced videos, we leverage 3D convolution layers to implement the denoising module. However, notably, it is also feasible to process each frame recursively with a denoising method of single image processing.

\textit{\textbf{Quadrilinear Interpolation.}}
Although we have presented a detailed definition of IA-LUT and its corresponding quadrilinear interpolation in this paper, the calculation formula for the offset $O_{\mathbf{x}}$ is omitted due to the space limitation. Therefore, we supplement the calculation formula here, which can be expressed as:
\begin{equation}
    O_{\mathbf{x}} = O_{r} \cdot O_{g} \cdot O_{b} \cdot O_{e},
\end{equation}
where we have the four terms
\begin{equation}
\begin{split}
    & O_{r} = \left\{ {\frac{r - r_{i}}{r_{i + 1} - r_{i}},\frac{r_{i + 1} - r}{r_{i + 1} - r_{i}}} \right\},~O_{g} = \left\{ {\frac{g - g_{i}}{g_{i + 1} - g_{i}},\frac{g_{i + 1} - g}{g_{i + 1} - g_{i}}} \right\}, \\
    & O_{b} = \left\{ {\frac{b - b_{i}}{b_{i + 1} - b_{i}},\frac{b_{i + 1} - b}{b_{i + 1} - b_{i}}} \right\},~O_{e} = \left\{ {\frac{e - e_{i}}{e_{i + 1} - e_{i}},\frac{e_{i + 1} - e}{e_{i + 1} - e_{i}}} \right\}, \\
\end{split}
\end{equation}
where $(r,g,b,e)$ denotes the input index, and $\mathbf{C}_{\mathbf{x}} = [r_{i}, g_{j}, b_{k}, e_{m}]$ stands for the index of grid point $\mathbf{x} = [i,j,k, m]$. All the terms appeared above have been defined in the paper. By utilizing this calculation formula, we are able to compute the 16 offsets required during the quadrilinear interpolation, respectively.

\textit{\textbf{Loss Function.}}
In addition to the loss between the output of our method and the ground truth, we also introduce 4D smooth regularization and 4D monotonicity regularization adapted to the IA-LUT into the loss function. The 4D smooth regularization is designed to ensure the stability of the conversion from the input space to the mapped color space, which helps avoid artifacts caused by extreme color changes in the IA-LUT. It consists of two parts which correspond to the video-adaptive IA-LUT and the video-dependent weights, respectively. 
Firstly, we have the 4D smooth regularization
\begin{equation}
    l_{s} = l_{lut} + l_{w},
\end{equation}
where $l_{lut}$ denotes the part related to the video-adaptive IA-LUT, and $l_{w}$ indicates the other part related to the video-dependent weights. For brevity, we here let
\begin{equation}
\begin{split}
    \mathbf{x}_i = \left\lbrack {i+1,j,k,m} \right\rbrack,~\mathbf{x}_j = \left\lbrack {i,j+1,k,m} \right\rbrack, \\
    \mathbf{x}_k = \left\lbrack {i,j,k+1,m} \right\rbrack,~\mathbf{x}_m = \left\lbrack {i,j,k,m+1} \right\rbrack, \\
\end{split}
\end{equation}
as the four grid points obtained by increasing the index of grid point $\mathbf{x} = \left\lbrack {i,j,k,m} \right\rbrack$ by one unit length along each of the four dimensions. Then, we define the function $\mathcal{F}$ as:
\begin{equation}
\begin{split}
\mathcal{F}\left( {i,j,k,m} \right) &= \left\| {C_{\mathbf{x}_i}^{'} - C_{\mathbf{x}}^{'}} \right\|^{2} + \left\| {C_{\mathbf{x}_j}^{'} - C_{\mathbf{x}}^{'}} \right\|^{2} \\
&+ \left\| {C_{\mathbf{x}_k}^{'} - C_{\mathbf{x}}^{'}} \right\|^{2} + \left\| {C_{\mathbf{x}_m}^{'} - C_{\mathbf{x}}^{'}} \right\|^{2}, \\
\end{split}
\end{equation}
where $\mathbf{C}'_{\mathbf{x}} = [r_{\mathbf{x}},g_{\mathbf{x}},b_{\mathbf{x}}]$ denotes the stored values in IA-LUT for grid point $\mathbf{x}$. Thus, we have
\begin{equation}
l_{s} = {\sum\limits_{i,j,k,m}{\mathcal{F}(i,j,k,m) + {\sum\limits_{n=1}^{T}\left\| w_{n} \right\|^{2}}}},
\end{equation}
where $w_{n}~,n \in \left\lbrack 1,2,...,T \right\rbrack$ denotes the $T$ video-dependent weights.

\begin{table*}[t]
  \centering
  \caption{Architecture of the encoder-decoder network, where "$nf$" is a hyper-parameter that serves as a channel multiplier controlling the width of each convolution layer. In this paper, the "$nf$" is set to 8.}
  \setlength{\tabcolsep}{5mm}{
    \begin{tabular}{ccccc}
    \toprule
    \multirow{2}{*}{Id} & \multicolumn{2}{c}{Encoder} & \multicolumn{2}{c}{Decoder} \\
\cmidrule{2-5}          & Layer & Output Shape & Layer & Output Shape \\
    \midrule
    0     & Conv $3\times3\times3$, Leaky ReLU & $nf\times H/2\times W/2$ & Deconv $3\times3\times3$, Leaky ReLU & $8nf\times H/32\times W/32$ \\
    1     & InstanceNorm & $nf\times H/2\times W/2$ & InstanceNorm & $8nf\times H/32\times W/32$ \\
    2     & Conv $3\times3\times3$, Leaky ReLU & $2nf\times H/4\times W/4$ & Deconv $3\times3\times3$, Leaky ReLU & $4nf\times H/8\times W/8$ \\
    3     & InstanceNorm & $2nf\times H/4\times W/4$ & InstanceNorm & $4nf\times H/8\times W/8$ \\
    4     & Conv $3\times3\times3$, Leaky ReLU & $4nf\times H/8\times W/8$ & Deconv $3\times3\times3$, Leaky ReLU & $2nf\times H/4\times W/4$ \\
    5     & InstanceNorm & $4nf\times H/8\times W/8$ & InstanceNorm & $2nf\times H/4\times W/4$ \\
    6     & Conv $3\times3\times3$, Leaky ReLU & $8nf\times H/16\times W/16$ & Deconv $3\times3\times3$, Leaky ReLU & $nf\times H/2\times W/2$ \\
    7     & InstanceNorm & $8nf\times H/16\times W/16$ & InstanceNorm & $nf\times H/2\times W/2$ \\
    8     & Conv $3\times3\times3$, Leaky ReLU & $8nf\times H/32\times W/32$ & Deconv $3\times3\times3$, Leaky ReLU & $1\times H\times W$ \\
    9     & InstanceNorm & $8nf\times H/32\times W/32$ & \textbackslash{} & \textbackslash{} \\
    \bottomrule
    \end{tabular}}%
  \label{tab:table1}%
  \vspace{-2mm}
\end{table*}%
As for the other regularization loss $l_{m}$, we let $\mathcal{M}(a)=max(a,0)$ and define the function $\mathcal{G}$ as:
\begin{equation}
\begin{split}
\mathcal{G}\left( {i,j,k,m} \right) &= \mathcal{M}\left( {r_{x_{i}} - r_{x}} \right) + \mathcal{M}\left( {g_{x_{i}} - g_{x}} \right) + \mathcal{M}\left( {b_{x_{i}} - b_{x}} \right) \\
&+ \mathcal{M}\left( {r_{x_{j}} - r_{x}} \right) + \mathcal{M}\left( {g_{x_{j}} - g_{x}} \right) + \mathcal{M}\left( {b_{x_{j}} - b_{x}} \right) \\
&+ \mathcal{M}\left( {r_{x_{k}} - r_{x}} \right) + \mathcal{M}\left( {g_{x_{k}} - g_{x}} \right) + \mathcal{M}\left( {b_{x_{k}} - b_{x}} \right) \\
&+ \mathcal{M}\left( {r_{x_{m}} - r_{x}} \right) + \mathcal{M}\left( {g_{x_{m}} - g_{x}} \right) + \mathcal{M}\left( {b_{x_{m}} - b_{x}} \right). \\
\end{split}
\end{equation}
So we can express the calculation of $l_{m}$ as:
\begin{equation}
    l_{m} = {\sum\limits_{i,j,k,m}{\mathcal{G}(i,j,k,m)}}.
\end{equation}

\section{Exploration of Hyper-Parameters}
\label{sec:experiment}
In terms of the number of grid points and the number of basis IA-LUTs, which determine the precision of the color transformation modeled by the generated LUT, their values cannot be increased indiscriminately due to the size limitation of IA-LUT. Although in this paper, we follow the most widely-used setting~\cite{zeng2020learning, yang2022adaint, yang2022seplut} of the two hyper-parameters $L$ and $T$, further experiments are needed to confirm the validity of this selected setting.

\begin{table}[t]
  \centering
  \caption{Quantitative results of FastLLVE with different numbers of $L$ on SDSD test dataset. The best is in \textcolor[rgb]{ 1,  0,  0}{red}.}
  \setlength{\tabcolsep}{4mm}{
    \begin{tabular}{c|cccc}
    \toprule
    L     & 9     & 17    & 33    & 64 \\
    \midrule
    PSNR  & 25.10 & 26.42 & \textcolor[rgb]{ 1,  0,  0}{\textbf{27.06}} & 24.52 \\
    \midrule
    SSIM  & 0.7717  & 0.7736  & \textcolor[rgb]{ 1,  0,  0}{\textbf{0.7769}}  & 0.7358 \\
    \bottomrule
    \end{tabular}}%
  \label{tab:table2}%
  \vspace{-4mm}
\end{table}%

To explore the influence of the number of grid points on the enhancement effect, as shown in Table~\ref{tab:table2}, we divide the IA-LUT into different numbers of unit lattices (i.e., different numbers of grid points). As expected, increasing $L$ from 9 to 33 improves the performance of our method. However, note that degraded performance is observed with the value of $L$ more than 33. One possible reason is that too high precision may result in over-fitting. Additionally, a large number of grid points also leads to heavy memory burden and great training difficulty which we should avoid. Therefore, with the consideration of avoiding over-fitting and reducing the size of IA-LUT, setting the number of grid points to 33 is indeed the one of the optimal choices.

\begin{table}[t]
  \centering
  \caption{Quantitative results of FastLLVE with different numbers of $T$ on SDSD test dataset. The best is in \textcolor[rgb]{ 1,  0,  0}{red}.}
  \setlength{\tabcolsep}{4mm}{
    \begin{tabular}{c|cccc}
    \toprule
    T     & 1     & 2     & 3     & 4 \\
    \midrule
    PSNR  & 26.40 & 26.58 & \textcolor[rgb]{ 1,  0,  0}{\textbf{27.06}} & 25.23 \\
    \midrule
    SSIM  & 0.7539  & 0.7648  & \textcolor[rgb]{ 1,  0,  0}{\textbf{0.7769}}  & 0.7641 \\
    \bottomrule
    \end{tabular}}%
  \label{tab:table3}%
  \vspace{-2mm}
\end{table}%

Similarly, to explore the influence of the number of basis IA-LUTs on the enhancement effect, as shown in Table~\ref{tab:table3}, we use different numbers of basis IA-LUTs to fuse the video-adaptive IA-LUTs. It can be observed that the performance of our method is positively correlated with the value of $T$. Then, similar to the above experiment, degraded performance appears when $T=4$, possibly due to the over-fitting. Besides, there is also a need for reducing the parameters of the fully connected layers. As a consequence, it may not be a good choice to further increase the number of basis IA-LUTs, compared with setting the value of $T$ to 3.

\begin{table}[t]
  \centering
  \caption{Quantitative results of FastLLVE+\emph{dn} with different numbers of $N$ on SDSD test dataset. The best is in \textcolor[rgb]{ 1,  0,  0}{red}.}
  \setlength{\tabcolsep}{4mm}{
    \begin{tabular}{c|cccc}
    \toprule
    T     & 5     & 7     & 9 \\
    \midrule
    PSNR  & 26.51 & \textcolor[rgb]{ 1,  0,  0}{\textbf{27.55}} & 26.12 \\
    \midrule
    SSIM  & 0.851 & \textcolor[rgb]{ 1,  0,  0}{\textbf{0.855}}  & 0.849 \\
    \midrule
    AB~(Var)↓  & \textcolor[rgb]{ 1,  0,  0}{\textbf{0.032}} & 0.033  & 0.052 \\
    \midrule
    MABD↓  & 0.049 & \textcolor[rgb]{ 1,  0,  0}{\textbf{0.040}}  & 0.055 \\
    \bottomrule
    \end{tabular}}%
  \label{tab:table4}%
  \vspace{-4mm}
\end{table}%

Finally, we consider the influence of the number of input frames on the enhancement effect. Therefore, we conduct an experiment, using different numbers of input frames, to investigate its effect on performance of the model. According to the Table~\ref{tab:table4}, our method achieves its optimal performance when $N$ is set to 7. Notably, increasing $N$ beyond this value leads to performance degradation, possibly attributable to challenges encountered by the model in achieving convergence. As a result, the experimental result suggests 7 input frames.

\end{document}